%% file: main.tex
\documentclass{article} 
\usepackage{collas2024_conference,times}

\input{math_commands.tex}

\usepackage{hyperref}
\hypersetup{
    colorlinks=true,
    linkcolor=red,
    filecolor=magenta,
    urlcolor=blue,
    citecolor=purple,
    pdftitle={Overleaf Example},
    pdfpagemode=FullScreen,
    }
\usepackage{url}
\usepackage{graphicx}
\usepackage{float}
\usepackage{soul}

\title{Keep Moving: identifying task-relevant subspaces to maximise plasticity for newly learned tasks
}

\author{Daniel Anthes\thanks{These authors contributed equally to this work.}\, , Sushrut Thorat$^{*}$, Peter König \& Tim C Kietzmann \\
Institute of Cognitive Science, Osnabrück University, Osnabrück, 49090, Germany \\
\texttt{\{danthes,sthorat,pkoenig,tkietzma\}@uos.de} \\}

\collasfinalcopy 

\begin{document}

\maketitle

\begin{abstract}
Continual learning algorithms strive to acquire new knowledge while preserving prior information. Often, these algorithms emphasise stability and restrict network updates upon learning new tasks. In many cases, such restrictions come at a cost to the model's plasticity, i.e. the model’s ability to adapt to the requirements of a new task. But is all change detrimental? Here, we approach this question by proposing that activation spaces in neural networks can be decomposed into two subspaces: a readout range in which change affects prior tasks and a null space in which change does not alter prior performance. Based on experiments with this novel technique, we show that, indeed, not all activation change is associated with forgetting. Instead, only change in the subspace visible to the readout of a task can lead to decreased stability, while restricting change outside of this subspace is associated only with a loss of plasticity. Analysing various commonly used algorithms, we show that regularisation-based techniques do not fully disentangle the two spaces and, as a result, restrict plasticity more than need be. We expand our results by investigating a linear model in which we can manipulate learning in the two subspaces directly and thus causally link activation changes to stability and plasticity. For hierarchical, nonlinear cases, we present an approximation that enables us to estimate functionally relevant subspaces at every layer of a deep nonlinear network, corroborating our previous insights. Together, this work provides novel means to derive insights into the mechanisms behind stability and plasticity in continual learning and may serve as a diagnostic tool to guide developments of future continual learning algorithms that stabilise inference while allowing maximal space for learning.\footnote{The code to replicate these results can be found at \url{https://github.com/KietzmannLab/keep-moving}}
\end{abstract}

\section{Introduction}

Catastrophic forgetting~\citep{french1999catastrophic, mccloskey1989catastrophic} is a key problem for continual learning. As networks are continuously trained to acquire new knowledge, performance on previously learned tasks rapidly decays. To avoid this problem, many algorithms exist that aim to stabilise networks as learning continues to novel task settings. Often, this is done by restricting learning with the underlying assumption that changes in the network are detrimental to stability and, necessarily, lead to forgetting. While this approach is generally successful at stabilising previously learned knowledge, it comes at the cost of limiting the network’s ability to adapt to new tasks - their plasticity. This problem is referred to as the stability-plasticity dilemma~\citep{carpenter1987massively, mermillod2013stability}. 

Yet, some results speak against an inevitable dilemma. Previous work has shown that changes in the activations of a network, as a result of learning new tasks, are not necessarily catastrophic. Even if, by behavioural measures, information is forgotten (i.e., classification performance decreases), information is often retained in the network and can be recovered with linear probes~\citep{davari2021probing, anthes2023diagnosing}. This opens the possibility that learning-induced activation changes are not inherently problematic. Rather, the ability of the network to change is a requirement for plasticity. What is thus needed is a way to disentangle change that affects previous performance (which affects stability) and change that does not (which allows continued plasticity). To test this hypothesis, activation change is decomposed into two orthogonal components that serve different roles regarding the stability and plasticity of the network. To arrive at this decomposition, we start at the task's readout (the last layer of the network, a linear classifier specific to the task) and split activation space into a subspace in which change affects the readout and the remaining space, which is invisible from the perspective of this readout. The larger this latter nullspace, the more flexible the network is to learn new tasks. This decomposition disentangles and maps the seemingly opposing demands of stability and plasticity onto two orthogonal subspaces. 

Based on this approach, our contributions are threefold: First, we introduce and utilise the above decomposition as an analysis tool to gain insight into existing algorithms and characterise their behaviour regarding stability and plasticity (Sections~\ref{sec:framework} and \ref{sec:comparison}).  Second, we study a simple linear system in which activation change can be directly manipulated in both subspaces. The respective magnitude of activation change can thus be causally linked to, respectively, stability and plasticity (Section~\ref{sec:linear}). Third, we present an approximate method that allows us to generalise our insights from the linear case and to deep nonlinear networks. In this context, additional complexities are discussed that arise from the nonlinear, hierarchical nature of deep networks (Section~\ref{sec:nonlinear}). Together, this work contributes to the understanding of how changes in neural networks during learning affect stability and plasticity. By disentangling the models' internal representational spaces, we demonstrate that nullspace movement during learning is beneficial to plasticity rather than detrimental, broadening the focus of diagnostic work in continual learning.

\subsection{Related Work}

While continual learning is commonly defined as learning in cases where data or tasks are non-stationary, several scenarios exist with differing assumptions about which aspects of the agent’s environment are non-stationary and what information about the environment is available to the learner~\citep{van2022three}. This work focuses on the task-incremental multi-head setup, where an agent is trained on multiple classification tasks sequentially, and each task comes with a separate dataset and corresponding labels. Thus, the learner has access to the `task label' at all times and can use a separate readout for each task. The remaining parameters of the network are shared across all tasks. 

Approaches to designing continual learning algorithms for the task-incremental setting can be clustered according to the information accessible by the agent throughout its lifetime. Consensus holds that three main groups exist: replay-based methods, regularisation methods, and architectural methods~\citep{de2021continual, parisi2019continual, hadsell2020embracing}. 
Here, we are interested in task-relevant subspaces in a shared network that sequentially learns multiple tasks without any changes to the network architecture. Therefore, we focus our analysis on algorithms of the first two groups: replay-based and regularisation methods that allow a  controlled analysis of their activation spaces. 

Besides the development of algorithms for continual learning, a body of diagnostic work exists that aims at characterising the behaviour of continually trained neural networks. 
This work has demonstrated that forgetting in continual learning concentrates in the last layers of a network and that the amount of forgetting is dependent on the similarity between new and old tasks~\citep{ramasesh2021effect, kalb2022causes}. Moreover, larger networks are less affected by forgetting, and pre-training additionally reduces the problem~\citep{ramasesh2020anatomy}. Yet, a focus on behaviour, i.e. quantifying forgetting as decreased classification performance, can lead to an overestimation of how much information is lost in a given network~\citep{davari2021probing}. While changes in activation patterns for previous tasks do lead to misalignment and decreased performance at task readouts, representational geometries and the discriminability of old classes can be largely preserved even if activations change~\citep{anthes2023diagnosing}. Finally, in addition to the field's focus on understanding forgetting and increasing stability, some previous work also started to look into plasticity, studying how it can decrease during continual learning~\citep{dohare2023loss}. Although a wealth of work on continual learning is available, a principled understanding of the stability-plasticity dilemma is still missing.  
    
Aside from studying stability and plasticity in isolation, recent work also addressed the stability-plasticity dilemma directly by investigating task-relevant subspaces~\citep{saha2021gradient, wang2021training, kong2022balancing, zhao2023rethinking, he2017overcoming}, proposing algorithms that estimate task-relevant subspaces by focusing on the activation spaces populated during previous tasks. 
However, estimating the task-relevant subspaces based on activation patterns has the potential to overestimate the space that must be preserved in order to achieve stability. The observation underlying this change of perspective is that networks can learn features that are orthogonal to the network’s task~\citep{hong2016explicit,thorat2021categoryorthogonal}.
As a result, the space that needs to be protected against change does not span the full space populated by the activations for a given task. That is, some subspaces encode information that is orthogonal to the task at hand. These spaces can be useful for future learning (plasticity) that will not interfere with existing input-output mapping. 
Consequently, instead of limiting the full subspace spanned by the activations of the previous tasks, previous work only limited the subspaces spanned by those activations that would have led to a loss of stability: 
\cite{he2018continual} shifts the focus from activation patterns at a specific layer to the effects on activation subspaces occupied at the next layer, limiting change to only subspaces occupied by a task's activations that are potentially propagated to the next layer. 
\cite{deng2021flattening} utilise replayed samples from old tasks to adaptively determine which directions in the subspace of old tasks most contribute to forgetting, allowing for changes in directions occupied by old activation patterns that least contribute to an increase in loss for old tasks.

Here, we adopt a similar, yet distinct view, by focusing only on subspaces where changes in activations are functionally relevant for a task through their downstream effect on the task’s readout. 
 In our work, we show that for a linear one layer network these subspaces can be derived directly from the weight matrix of the last layer, while in deep nonlinear networks they can be estimated using the task's gradients similar to \cite{farajtabar2020orthogonal}. We use these methods as a diagnostic tool that separates task-relevant and task-orthogonal subspaces and relates movement therein to effects on the stability and plasticity of continual learners.

\section{Identifying task-relevant subspaces and task-orthogonal nullspaces to diagnose continual learners}
\label{sec:framework}

\begin{figure}[!t]
\centering
\includegraphics[width=\textwidth]{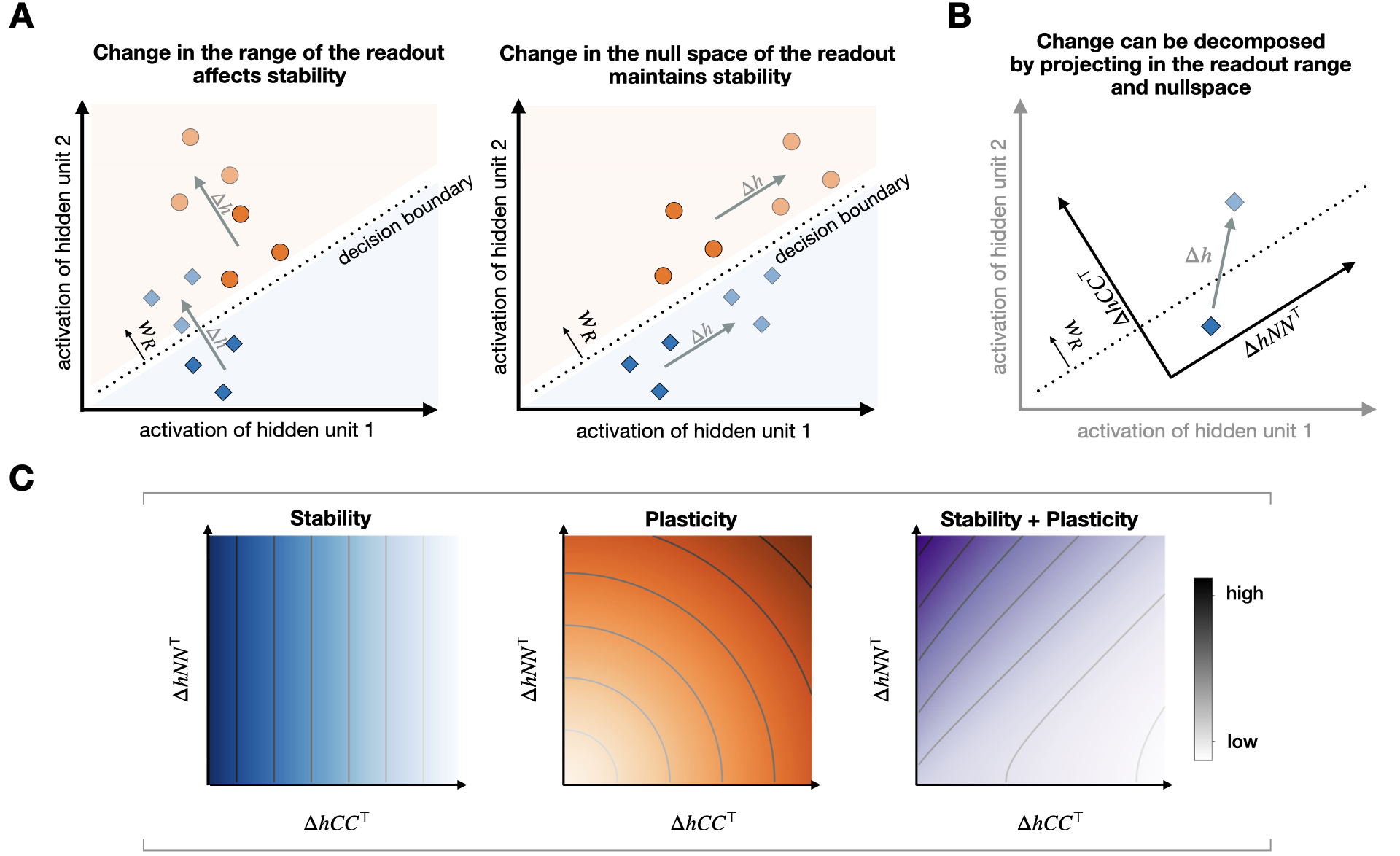}
\label{fig:conceptual}
\caption{\textbf{Linking stability and plasticity to changes in activations seen by prior readouts.} \textbf{(A)} Learning for a new task can cause two kinds of activation change from the perspective of the old task's readout.
Changes perpendicular to the decision boundary for the old task can affect stability (left).
Changes parallel to the decision boundary are invisible to the readout and cannot cause forgetting (right).
The conceptual plots show the change in hypothetical activation patterns. 
\textbf{(B)} The readout range and nullspace define a new basis in which activation change can be meaningfully linked to stability and plasticity, respectively. \textbf{(C)} Activation change in the range of the old readout $\left(\mathbf{CC^{\top}}\right)$ affects stability while restricting activation space in any subspace may be detrimental to plasticity. Taking into account both constraints, a successful continual learner is expected to restrict only learning in the range of old tasks.}
\end{figure}

The pre-readout layer is responsible for the largest amount of change in the network \citep{ramasesh2020anatomy, kalb2022causes}, and its activation changes summarise all changes that can affect the readout. Therefore, we argue that studying activation change at the last hidden layer of the network can be informative to diagnose how learning in all layers of the network affects performance for previously learned tasks.
Here, we decompose the changes of activation in this layer into two components that play functionally distinct roles during learning: movement in the readout range of a network affects previous task performance, while movement in the orthogonal nullspace does not. Thus, changes in the former should be constrained to achieve stability, and changes in the latter should be constrained as little as possible to maintain plasticity (Fig.~\ref{fig:conceptual}A).

\subsection{Notation and definition of relevant subspace} 

In task-incremental continual learning, a network has to sequentially learn a set of $n$ task mappings \{$\mathbf{x^k\rightarrow o^k}$\}$_{k\leq n}$, mapping the samples $\mathbf{x^k}$ for task  k to predicted outputs $\mathbf{o^k}$.
The network's weights are shared across tasks, except for the parameters of the last layer (hereafter referred to as 'readout layer'), which are specific to each task $k$. 
Since all but the last layer of the network are shared for all tasks, we can write the function learned for task $k$  as $ \mathbf{o^k = r_k(f(x^k)) }$ . Here, $\mathbf{f(x^k) = h^k}$ is the model backbone shared by all tasks, mapping the samples for a task to activations $\mathbf{h^k} \in \mathbb{R}^{1 \times p}$ at the last shared hidden layer, and $\mathbf{r_k(h^k) = o^k}$ is the task-specific readout layer which maps the hidden activations to output probabilities $\mathbf{o^k} \in \mathbb{R}^{ 1 \times q}$ for all q classes included in task $k$.
For a given task $k$ consider the activations $\mathbf{h^k}$ at the final hidden layer and the task readout layer $\mathbf{r_k(h^k)}$ such that $\mathbf{o^k = r_k(h^k) = \sigma(h^k W_{R^k}^{\top}+b_{R^k}^{\top})}$.
Here, $\mathbf{W_{R^k}} \in \mathbb{R}^{q \times p}$ and $\mathbf{b_{R^k}} \in \mathbb{R}^{q \times 1}$ are the weight matrix and bias vector learned for task $k$, and $\sigma$ is a non-linear activation function, usually a softmax operation mapping activity of the output units to class probabilities\footnote{However, to maintain a stable input-output mapping for a task, it is important to consider a further, crucial, nonlinearity that is the Argmax operation, mapping logits to predicted labels.}. 
While learning the new task $m$ ($m>k$), at each step of gradient descent, the hidden activations $\mathbf{h^k}$ for the old task's data $\mathbf{x^k}$ change as a result of updates to the weights in the shared backbone of the network.
The weights of the readout for this task $\mathbf{W_{R^k}}$, on the other hand, remain fixed, as no gradients flow through old readouts during training for a new task.

The readout weight matrix $\mathbf{W_{R^k}}$  consists of one row vector $\mathbf{w_{{R^k}_i}}$ per output unit $o^{k}_i$ of task $k$. This vector determines the direction in the activation space of the last hidden layer that this readout unit is sensitive to (indicated with a vector labeled $\mathbf{W_R}$ in Fig.\ref{fig:conceptual}). Jointly, the directions for all $o^k_i$ span the subspace that is visible to the task's readout. This subspace is called the range of $\mathbf{W_{R^k}}$ and can be obtained by finding the orthonormal basis $\mathbf{C} \in \mathbb{R}^{p \times m}$ of the readout weight matrix using singular value decomposition (SVD)\footnote{In the case where the dimensionality of the pre-readout layer is larger than the number of classes in a task (which is usually the case), the number of nonzero singular values is bounded from above by the number of classes in task $k$. In the following analyses utilising this decomposition of  $\mathbf{W_{R^k}^{\top}}$ we keep all dimensions with nonzero singular values when computing $\mathbf{C}$, but the effective dimensionality may be further reduced by discarding directions with small singular values if the readout directions for different classes are strongly correlated.}. The remaining directions in the activation space of the last hidden layer, orthogonal to the range, are called the nullspace of the matrix $\mathbf{N} \in \mathbb{R}^{p \times (p - m)}$. Together, these two subspaces span the entire activation space that $\mathbf{W_{R^k}}$ operates on.

This property is used to decompose the change in activations $\mathbf{\Delta h^k}$ as $\mathbf{\Delta h^k CC^{\top}+ \Delta h^k NN^{\top}}$ (Fig.~\ref{fig:conceptual}B). 
Here, $\mathbf{C}$ and $\mathbf{N}$ are the range and nullspace matrices for task $k$ obtained by decomposing $\mathbf{W_{R^k}}$ as described above, 
and the two matrices $\mathbf{CC^{\top}}$ and $\mathbf{NN^{\top}}$ are the corresponding projection matrices that divide the pre-readout layer activation space into two orthogonal components (see Fig. \ref{fig:conceptual}B). 
These two components play distinct roles in the stability of the task mapping $x^k \to o^k$.

Changes in the activations in the range of the readout ($\mathbf{\Delta h^k CC^{\top}}$) can potentially disrupt stability as $\mathbf{\Delta h^k CC^{\top} W_{R^k}^{\top} = \Delta h^k W_{R^k}^{\top}}$, meaning that changes in this space are visible to the readout.
In contrast, activation changes in the nullspace of the readout ($\mathbf{\Delta h^k NN^{\top}}$) are invisible to the task readout, as by definition $\mathbf{NN^{\top} W_{R^k}^{\top} = 0}$ and, as a result, $\mathbf{\Delta h^k NN^{\top} W_{R^k}^{\top}=0}$.
The actual functionally relevant subspace may not take up the entire range of the readout weights due to the readout nonlinearity and data statistics. Still, all functionally relevant dimensions for the old task must be contained in this range. Therefore, the range of the readout defines an 'upper bound' on the space in which changes can affect stability. There is no benefit for stability in restricting the null space of old readouts.
Restricting the space in which learning for the new task is allowed without disturbing the stability of previous tasks could, however, reduce plasticity.

\section{Result 1: Existing regularisation algorithms are overconstrained by restricting activation change in the nullspace}
\label{sec:comparison}

\begin{figure}[!t]
\centering
\includegraphics[width=\textwidth]{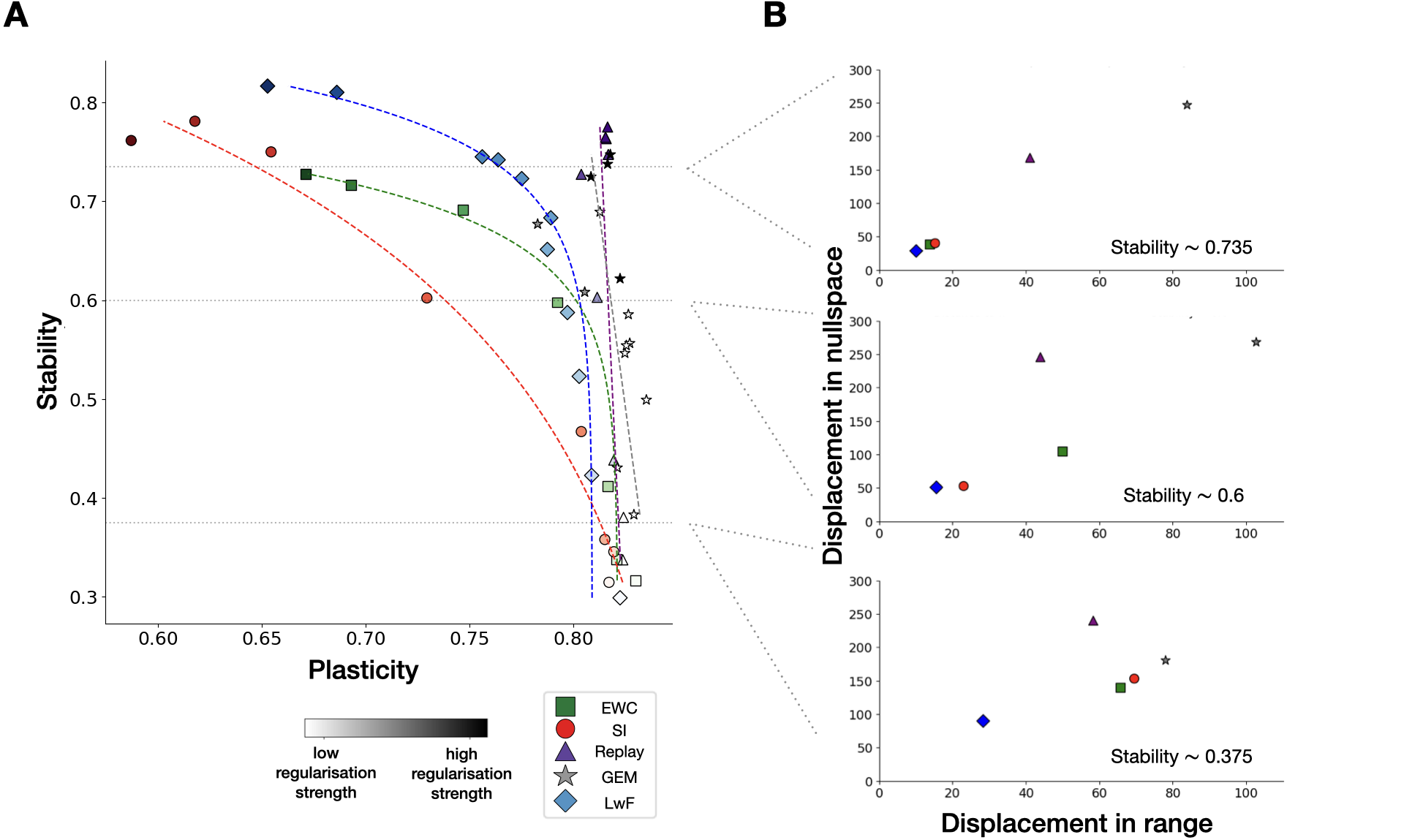}
\label{fig:comparison}
\caption{
\textbf{Stability and plasticity trade-off curves comparing selected continual learning algorithms} 
\textbf{(A)} Stability and plasticity of selected regularisation and replay methods for continual learning. Hue indicates the algorithm used and shading indicates the strength of regularisation used (For a full list of parameters for each algorithm see \ref{sec:appendix_nonlinear}). Dotted lines indicate three levels of stability for which we compare activation change in panel B. With increasing regularisation strength, the replay-based algorithms - data replay and GEM - maintain higher plasticity while maintaining high stability, as compared to the regularisation algorithms - EWC, SI, and LwF.
\textbf{(B)} Activation change at the pre-readout layer for data from the first task as a result of learning $10$ additional tasks are shown. Activation change is decomposed into the range and null space of the readout for task 1. The three panels show the activation change for the tested algorithms, approximately matched for stability (at the stability levels indicated in panel A with dotted lines). At a given stability level, a higher degree of activation change in null space corresponds to more plasticity. Analyses of the stability-plasticity trade-offs of these algorithms and the corresponding displacements in range and null space of task 1 readout are shown in Appendix Figure \ref{fig:comparison_over_phases}.}

\end{figure}

The above decomposition of activation change is applied to analyse the behaviour of a set of benchmark continual learning algorithms in a well-known task and network architecture. 
Benchmark algorithms include two regularization methods - Synaptic Intelligence (SI; \citet{zenke2017continual}) and EWC \cite{kirkpatrick2017overcoming}, and three rehearsal methods - Learning without Forgetting (LwF; \citet{li2017learning}), Gradient Episodic Memory (GEM; \citet{lopez2017gradient}) and data replay~\citep{rebuffi2017icarl}\footnote{In data replay and LwF, contrary to the usual setting, readouts are fixed after training on a task and allowed the rest of the network to train further. This was done because, in all the other methods, the prior readouts are frozen. Avalanche library was used for the implementation~\citep{carta2023avalanche}}.

EWC~\citep{kirkpatrick2017overcoming} and SI~\citep{zenke2017continual} restrain the learning trajectory by estimating the importance of the network's parameters during learning and placing a penalty on changes to these parameters during learning for subsequent tasks. The strength of the penalty for changing a parameter is proportional to its estimated importance for previous tasks.

Data replay~\citep{rebuffi2017icarl, bagus2021investigation} keeps a buffer with a subset of the data for old tasks and uses these data for joint optimisation of the old and new tasks. As a representative of this class, GEM~\citep{lopez2017gradient} keeps a replay buffer but uses the gradient for old tasks to project gradients for the new task in a direction that does not increase the loss for old tasks. This is achieved through an inequality constraint that enforces positive cosine similarity between the replayed and new task gradient. LwF~\citep{li2017learning} provides an interesting hybrid case. While this method keeps a replay buffer, this buffer is filled by recording the activations at old task readouts for data from the new task before the start of learning. 
This procedure creates a dataset of 'pseudo labels' based on which the algorithm strives to preserve learned input-output mappings at the readouts for previous tasks. 
Commonly, this algorithm is grouped with SI and EWC as a regularisation method~\citep{de2021continual}. However, in this work, the important distinction between methods is whether they can evaluate the loss function for previous tasks. Therefore, we group LwF as a replay method, even though LwF, as opposed to 'true' replay methods, can evaluate the loss landscape for old tasks only in the span of the dataset for the new task. 

\subsection{Network, Task, and Training}

Our experiments are based on a VGG-style neural network, similar to the one used in \citet{zenke2017continual}. The network maps CIFAR images~\citep{krizhevsky2009learning} to their classes (
Appendix~\ref{sec:appendix_nonlinear}). 
The networks are trained on the CIFAR-110 continual learning benchmark, in which the network is first trained on CIFAR-10, and then sequentially trained on ten equal task splits from CIFAR-100.
The analysis was repeated $3$ times. Average results are reported. Images were subjected to augmentations, and readouts were trained for each of the tasks, with softmax activation and cross-entropy loss. 
After training on the first task, the projection matrices $\mathbf{CC^{\top}}$ and $\mathbf{NN^{\top}}$ are computed to assess how further learning affects the range and nullspace of the first task's readout. 
To assess the behaviour of the tested algorithms, `stability' is defined as classification accuracy on the validation set of the first task after training on all $11$ tasks, and `plasticity' as the accuracy on the validation set of the last task.
For all algorithms, we systematically vary the respective hyperparameters and observe the resulting behavioural effects on the networks' stability and plasticity. In parallel, we decompose activation changes in the final hidden layer, linking the stability and plasticity of the algorithms to activation changes in range and nullspace of the readout for the first task.


\subsection{Results \& Interpretation}

First, we compare the stability and plasticity of the set of algorithms. With increasing regularisation strength, all tested algorithms successfully stabilise old knowledge in the network. This observation is especially true for the regularisation algorithms, EWC and SI, which trade off plasticity in the parameter regime to achieve high stability. This effect is more pronounced in SI. However, both algorithms are outperformed by the three replay-based methods; data replay and GEM, which do not lose any plasticity even in the regime of very high stability. LwF, on the other hand, does sacrifice some plasticity in cases where stability is very high. Thus, for matching levels of stability, we see that the algorithms differ in their ability to maintain plasticity (Fig.~\ref{fig:comparison}A).

To link the stability and plasticity of these algorithms to how they constrain activation change, we investigate how far activations at the pre-readout layer are displaced for data from the first task by continual learning. This displacement is computed as the Euclidean distance between activation patterns at the pre-readout layer after learning the first task initially, compared to the activation patterns for the same data after training on $10$ additional tasks. Importantly, the distances are computed separately for the two subspaces, given by the range and nullspace of the first task readout.

As stability increases, EWC, SI, and LwF start to constrain activation changes in both subspaces. The decrease in activation change correlates with the observed loss of plasticity. This suggests that to preserve input-output mappings for previously learned tasks, these algorithms (perhaps unnecessarily) restrict learning in substantial parts of the network's representational space (Fig.~\ref{fig:comparison}B). 
As stability is increased further, movement in both the range and nullspace is decreased. This suggests that the regularisation algorithms cannot fully disentangle learning in the range and nullspace and, as a result, end up restricting both.

Interestingly, GEM and data replay do not obey the same pattern and can achieve high stability despite continuing to move in the range and nullspace of previous readouts.
This behaviour may be the result of the algorithm's access to data from previous tasks and the resulting ability to evaluate the loss landscape for these tasks. 
The additional information allows the algorithms to make changes to the range of previous tasks that either do not disturb their input-output mappings or even improve them (i.e. backward transfer).
Despite its access to the loss landscape of previous tasks through `pseudo replay', LwF significantly restricts activation change and loses relatively more plasticity compared to the other replay methods. 
One possible explanation for this difference is that access to old data allows replay methods to know which parts of the readout range are `actually' occupied by the data of this task. 
Learning in parts of the range that are not occupied by the task's data is safe even if these changes are theoretically visible to old readouts. 
In LwF, on the other hand, the span of the pseudo replay data and new data is by definition equivalent, and as a result, LwF is the only method where constraints on learning are computed specifically for the subspace spanned by the data for the new task (even the importances computed by EWC and SI are computed in the span of the data of the old task).

To summarise, while existing regularisation methods succeed in preserving stability by limiting task-relevant activation change for old tasks, 
they do not fully separate task-relevant and irrelevant spaces. As a result, when regularisation is strong, the algorithms over-constrain the model and thereby reduce the potential for plasticity.
Algorithms, where information about the data and loss landscape for previous tasks are available during learning new tasks, have access to more efficient constraints that allow learning in the range of previous tasks without sacrificing stability. 
This increased freedom manifests in more activation change for previous tasks, which in turn results in increased plasticity. We find that these patterns generalise to a larger network and dataset (a ResNet trained on Mini Imagenet, see Appendix \ref{sec:appendix_tinyimg}).


\begin{figure}[!t]
\centering
\includegraphics[width=\textwidth]
{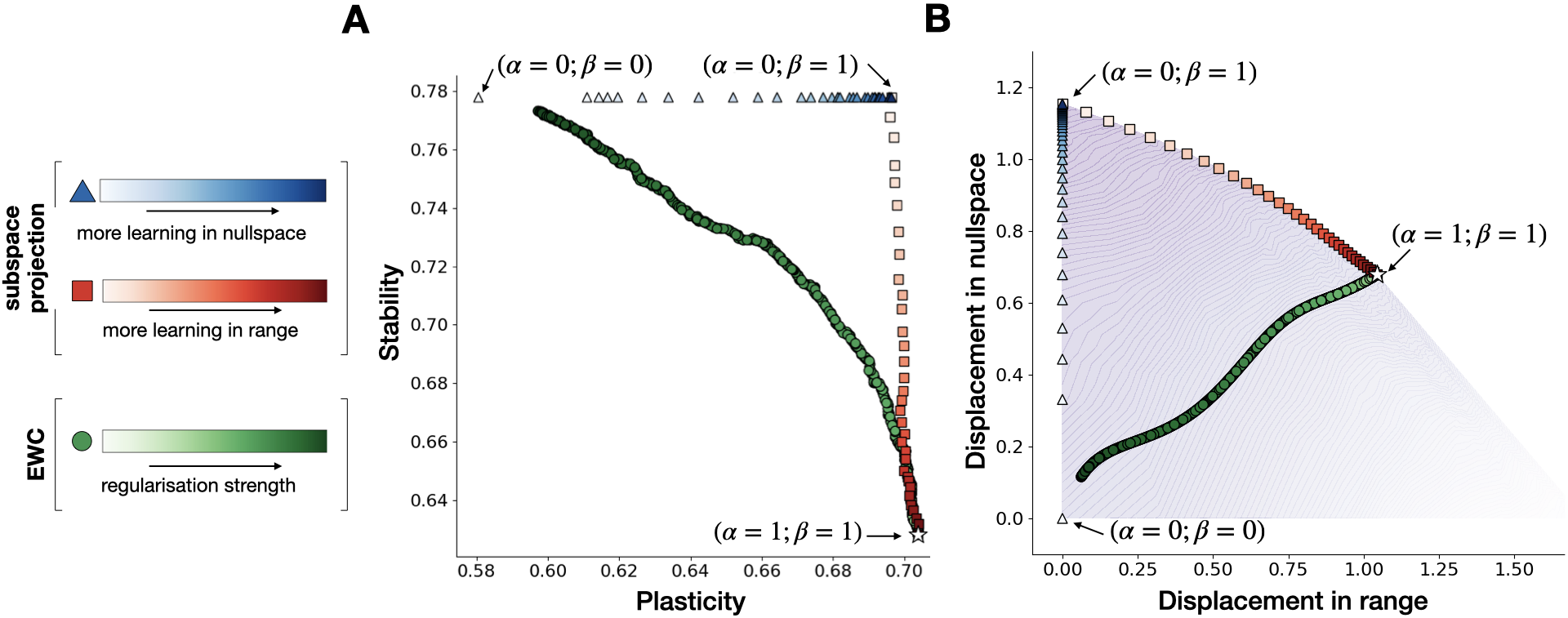}
\label{fig:linear}
\caption{\textbf{Plasticity and stability achieved by a one hidden-layer linear neural network trained on the Split MNIST task with gradient decomposition in the range and nullspace of the old task readout.}
\textbf{(A)} Plasticity and stability of the network trained with different configurations for $\alpha$ and $\beta$. 
Stability and plasticity of networks trained with gradient-based subspace decomposition and EWC. Each data point shows the performance of a network on the first task (stability) and the second task (plasticity), after training on both tasks. Data points are coloured according to the algorithm and parameters used. Green hues indicate networks trained with EWC with darker shades indicating stronger regularisation. Red and blue hues indicate networks trained with readout weight-based activation decomposition into the old readout's range and nullspace. Red hues show networks where only learning in the range is restricted ($\alpha$ is varied, $\beta=1$). Blue hues indicate  networks where learning in the range is restricted completely ($\alpha=0$) and restrictions on the functional nullspace are varied ($\beta$). \textbf{(B)} Activation change for data of the first task as a result of learning the second task. Data points show movement corresponding to stability and plasticity results in (A). The color of the overlaid contour indicates Stability + Plasticity (as in Fig.\ref{fig:conceptual}C) A darker colour indicates high stability and plasticity. For extended results see Fig.~\ref{fig:extended_contour}}
\end{figure}

\section{Result 2: Restricting learning to the nullspace achieves optimal stability and plasticity in a linear one-hidden layer network} 
\label{sec:linear}

The previous section has shown that separating activation change into two subspaces can yield novel insights into existing methods and their stability-plasticity tradeoff. Next, we utilize the decomposition of the activation space into the task readout's range and nullspace to directly control the degree to which either space is allowed to change during the learning of a new task.  This allows us to more causally assess the contribution of learning in these spaces to the stability and plasticity of a network.  We start by considering a simple linear network with one hidden layer and describe how, in this setting, our subspace decomposition based on the readout weight matrix can be used to control learning for a new task directly.

\subsection{Setup}

The input-output mapping of a one-hidden-layer linear network with no biases can be written as $\mathbf{o = xW_H^{\top}W_R^{\top}}$, where $\mathbf{o}^{1\times o}$ is the output with $o$ neurons, and $\mathbf{x}^{1\times x}$ is the input with $x$ features.
$\mathbf{W}_\mathbf{H}^{h\times x}$ is the mapping from the input to the hidden layer with $h$ neurons, and $\mathbf{W}_\mathbf{R}^{o\times h}$ is the mapping from the hidden layer to the output. 
After learning the first task, the network has learned the input-output mapping $\{\mathbf{x^1\rightarrow o^1}\}$ based on the trained weights for the hidden layer $\mathbf{W_H}$  and readout $\mathbf{W_{R^1}}$. 
While learning a new mapping for the data of the second task $\{\mathbf{x^2\rightarrow o^2}\}$, the gradient $\mathbf{\Delta\!W_H}$ for the shared hidden layer potentially causes activation changes at the hidden layer that affect the learned mapping for the first task. To maintain stability, we want to constrain learning in ($\mathbf{W_H}$) such that the learned mapping $\{\mathbf{x^1\rightarrow o^1}\}$ for task 1 is preserved. 

Learning can be constrained with a projection matrix $\mathbf{A}^{h\times h}$ applied to the gradient.  To avoid forgetting, we want the projected gradients $\mathbf{A\Delta\!W_H}$ to change the weights of the hidden layer such that the mapping for the first task is unchanged: 
\begin{equation}
\label{eq:RDAC_eq}
\mathbf{o^1 = x^1(W_H+A\Delta\!W_H)^{\top}W_{R^1}^{\top} \implies x^1(A\Delta\!W_H)^{\top}W_{R^1}^{\top}=0 \implies W_{R^1}A=0}
\end{equation}

One solution to this equation is to choose $\textbf{A}$ such that $\mathbf{A=NN^{\top}}$, where $\mathbf{N}$ is the nullspace matrix of $\mathbf{W_{R^1}}$. Projecting the gradient for  $\mathbf{W_H}$ into the nullspace of $\mathbf{W_{R^1}}$ is therefore guaranteed to preserve stability. Meanwhile, the hidden layer activations $\mathbf{h^1}$ are free to change in the nullspace of $\mathbf{W_{R^1}}$ which allows for plasticity. 

To allow full control over learning in the two subspaces, the gradients in either subspace are weighted with two scalar hyperparameters that control the amount of change in the range and nullspace of the previous task: $\mathbf{A = \alpha CC^{\top}+ \beta NN^{\top}}$.  $\alpha$ is the amount of learning allowed in the range, and $\beta$ is the amount of learning allowed in the nullspace. $\mathbf{W_{R^1}A=0}$ only if no learning is allowed in the range ($\alpha=0$), which ensures stability. Stability does not rely on the nullspace weight $\beta$ but plasticity does as, if $\beta$ is reduced while $\alpha=0$, the gradient becomes smaller, and learning the new task becomes slower.

To assess how stability and plasticity are affected by learning in the range and nullspace of the first task we apply our gradient projection method and vary the hyperparameters $\alpha$ and $\beta$. A linear model with one hidden layer with $11$ units is trained to classify MNIST digits. Training is split into two tasks, containing data for digits 0-4 and 5-9 respectively. 

\subsection{Results \& Interpretation}

Progressively restricting learning to the null space of the first task by decreasing $\alpha$, increases the stability of the model until maximal stability is reached. This is achieved when learning is fully restricted to the null space. In this setting, the model has similar stability to a model where the hidden weights are frozen completely (indicated in Fig.~\ref{fig:linear} as $\alpha = 0; \beta = 0$).
Although the capacity of our model is very small, restricting learning to the nullspace of the previous task allows the model to lose no plasticity compared to the unregularised model (indicated in fig~\ref{fig:linear} as $\alpha = 1; \beta = 1$).
Further restricting learning in the null space, in addition to the range, decreases plasticity.
Together these findings reinforce our intuition that the range of a previous task is singularly important for stability while restricting the nullspace hampers plasticity with no effect on stability.

As an additional point of comparison, the same linear network is also trained using EWC with varying regularisation strength. In line with our findings from the comparative analysis, increasing regularisation strength in EWC restricts learning in both the range and null space of the previous task. As a result, the EWC regularised network trades off plasticity for stability by over-constraining the null space of the first task.

The analysis presented here has the desirable property that the computation of the functionally relevant subspace for the first task depends only on the weights of the old readout. This conveys the intuition that all the information about which spaces are functionally relevant to a task has been absorbed in the network weights and can be recovered without any dependence on the task's data. Additionally, the old task's readout does not change during learning for a new task and therefore the decomposition only has to be computed once.

\begin{figure}[!t]
    \centering
    \includegraphics[width=\textwidth]{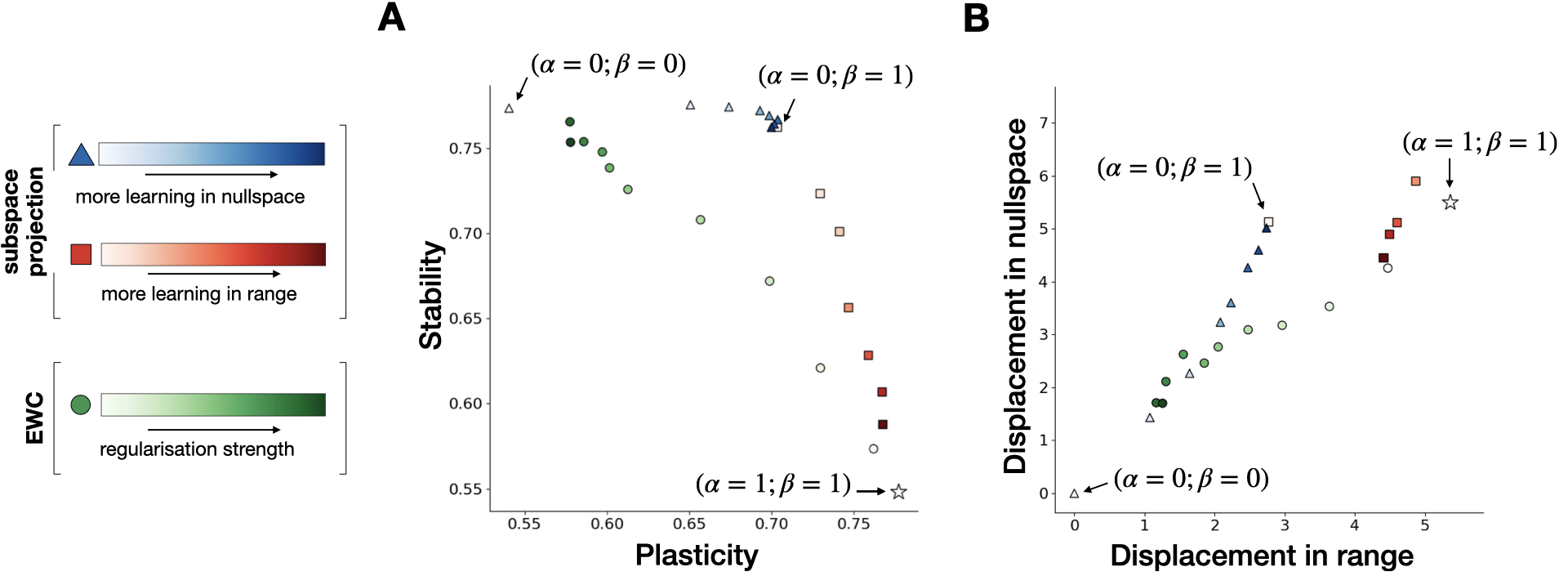}
    \caption{\textbf{Stability and plasticity in a nonlinear network trained on Split CIFAR-10 using gradient-based activation decomposition and EWC.} \textbf{(A)} Stability and Plasticity of networks trained with gradient-based subspace decomposition and with EWC. Each data point shows the performance of a network on the first task (stability) and the second task (plasticity), after training on both tasks. Data points are coloured according to the algorithm and parameters used. Green hues indicate networks trained with EWC with darker shades indicating stronger regularisation. Red and blue hues indicate networks trained with gradient-based activation decomposition. Red hues show networks where only the functional range is restricted ($\alpha$ is varied, $\beta=1$ ). Blue hues indicate networks where learning in the range is restricted completely ($\alpha=0$) and restrictions on the functional nullspace are varied ($\beta$).
    \textbf{(B)} Activation change at the pre-readout layer for data from the first task as a result of learning the second task. Activation change is decomposed into the change in the range of the first task's readout (Displacement in range, $CC^{\top}$) and change in its nullspace (Displacement in nullspace, $NN^{\top}$).
    In both panels points of interest are labelled: the baseline condition, where learning is unrestricted ($\alpha=1, \beta=1$), the condition where learning is restricted completely to the functional nullspace ($\alpha=0, \beta=1$), and the condition where the model's hidden layers are fully frozen ($\alpha=0, \beta=0$).}
    \label{fig:nonlinear}
\end{figure}

\section{Result 3: In deep nonlinear networks, the space spanned by the gradients of a task is an estimate of its functional range}
\label{sec:nonlinear}

\subsection{Notation and Setup}

Unfortunately, the analytical decomposition of activation spaces becomes nontrivial in more complex networks: 
with nonlinear activation functions, the projection matrix computed on the readout weights no longer acts directly on the weights of the hidden layer and cannot be used directly to filter the gradient updates $\mathbf{\Delta W_H}$.  Similarly, adding multiple hidden layers to the model causes the allowed updates to a layer to depend on the updates of all upstream layers, making the analytical identification of the nullspaces complicated, even for linear deep networks (see Appendix~\ref{sec:appendix_3layer}).
Instead, in this section, we discuss an approximation that allows for the estimation of this space at every layer of the network and can be computed for networks with multiple hidden layers and nonlinearities. This approximation takes into account the functional relevance of  changes in all layers of a network and allows us to generalise our analytical results to a wide range of nonlinear systems. We use these estimated spaces to verify that the findings from the analysis of the single hidden layer linear case generalise to a deep nonlinear network trained on the Split CIFAR-10 benchmark.

To extend the concept of the range and nullspace of a task's readout to a deep nonlinear network, we need to be able to trace the subspace visible to the readout through all layers in the network. Subsequently, projecting the gradient into the orthogonal, null, space at every layer is expected to restrict learning such that the previously learned input-output mapping remains unchanged \citep{farajtabar2020orthogonal}. 
To distinguish the functionally relevant space in a multi-layer non-linear network from the range of the readout matrix (as used in the decomposition of the readout weight matrix), we call the task-relevant space the `functional range' of a task, and the corresponding nullspace the `functional nullspace'.

To trace the subspace visible to the readout throughout the network we make use of a similar procedure as described in \cite{farajtabar2020orthogonal}: the gradients for prior tasks, given the corresponding prior readouts, signal the directions of weight change that affect the prior task mappings. Projecting new gradients into the null space of the prior task gradients would ensure the prior task mappings do not change, i.e. maintain stability, while allowing learning new tasks i.e. allow plasticity. In contrast to \cite{farajtabar2020orthogonal} however, we approximate the 'functionally relevant subspace' using the gradient of the old task's loss as opposed to the model gradients since we are interested in a tight estimate of the functionally relevant dimensions. We reason that the gradient of the loss allows for a smaller space to be restricted since we can exploit degeneracies introduced by the softmax nonlinearity. \footnote{However, see \cite{farajtabar2020orthogonal} for a discussion of why utilising the model gradients may yield better estimates for designing an algorithm.}

To approximate these spaces, all gradients are computed based on the intuition that mini-batches for a task's data must lie in a subspace that is visible to the task's readout. Gradients always lie in the span of the data for a task~\citep{saha2021gradient, zhang2021understanding}, and weight changes in the space of the old task's gradients cause activation changes in the functional range of the old task. Hence these changes affect the old task's input-output mapping.
The directions orthogonal to the subspace spanned by the old task's gradients constitute the old task's functional nullspace and cannot affect its input-output mapping. Projecting gradients for the new task in this space should therefore lead to learning without affecting stability \citep{farajtabar2020orthogonal}.

Additional batches of data are passed through the model after the training for a task has finished, and the gradients for the weights are computed at every layer of the network.
The span of the sampled gradients at a layer gives an approximation of the functional range of the task at this layer.  The subspace spanned by these gradients can be computed using SVD.
The orthonormal basis of the matrix of sampled gradients is the `functional range' at this layer, and the complementary dimensions are the `functional nullspace'. 
As opposed to the weight-based decomposition discussed in the context of the linear case, the estimates of functional range and functional nullspace here depend on the data of the previous task and are, therefore, useful primarily as an analysis tool in situations where we can access data for all tasks. Additionally, these linear estimates of the subspaces are computed for specific locations in the model's weight space and can become inaccurate after taking many steps in weight space \citep{farajtabar2020orthogonal}. We therefore frequently re-estimate them during training for the new task to achieve more accurate constraints on the gradients for the new task.

Using these estimates, the analysis described for the linear case is repeated in the nonlinear setting.
This analysis is performed on a deep convolutional network trained on the Split CIFAR-10 benchmark, where the network is sequentially trained on two distinct subsets of the CIFAR-10 dataset consisting of $5$ classes each.
For a detailed description of the experiment please refer to appendix \ref{sec_appendix_nonlinear_analysis}.

\subsection{Results \& interpretation}

Analogously to the linear case, projecting the gradients for the new task into the functional nullspace of the previous task stabilises the network almost completely. The model's plasticity slightly decreases when restricting learning to the functional nullspace of the model. However, additional constraints on the functional nullspace of the model have a far greater effect on plasticity (see Fig.~\ref{fig:nonlinear}). 
Compared to networks trained with EWC, projecting the gradients for the new task in the functional nullspace of the first task achieves greater stability and plasticity, in line with observations from the linear analysis.

Restricting learning for the new task in the functional nullspace of the first task greatly reduces activation change in the range of the old readout but not its nullspace. Contrary to the linear case (section \ref{sec:linear}), the transition from the unconstrained network ($\alpha=1; \beta=1$) to the network where learning is restricted fully to the functional nullspace does not lead to a smooth decrease of activation change in the range of the first task's readout.
One possible explanation is that the loss landscape for deep networks is not convex, and the algorithm can find solutions that exploit the functional range of the previous task even if the learning rate in this subspace is greatly reduced. Completely restricting learning to the functional nullspace, however, has a similar effect on activation change as in the linear case.

Activation change in the range of the first task's readout is not stopped completely if learning is restricted to the functional nullspace. This is explained by the estimate of the functional range being based on the gradients of the old task. These estimates take into account both the span of activations at the pre-readout layer and the nonlinearities connecting activations at the readout with the loss function.
Therefore, the functional nullspace is likely smaller than the subspace spanned by the readout weights.

Restricting learning in the functional nullspace of the previous task too, further decreases activation movement. As in the linear case, this activation change reduction correlates with significant plasticity decreases (Fig \ref{fig:nonlinear}).

As in previous analyses, increasing the regularisation strength of EWC decreases activation change in both the functional range and functional nullspace of the first task suggesting that the lower performance of EWC is explained by its over-constraining of the network.

To conclude this section, the main observations from the linear analysis generalise to the nonlinear case. Additionally, we found that the functionally relevant subspace for a task is even smaller than the space spanned by the readout's range. However, computing this space is much more expensive and depends on data for the task whose range we wish to estimate. Hence, the decomposition based on the task readout's weights remains a useful tool for characterising the behaviour of continual learners.

\section{Conclusion}

In this study, we presented an analysis of the conditions for stability and plasticity in continual learners with a primary focus on learning in the scenario where the learner cannot re-evaluate the loss landscape for previous tasks, i.e., by employing data replay. We show that in this scenario, stability can be achieved by restricting learning to the functional nullspace of previous tasks, while maintaining plasticity. We discuss the generality of our findings and avenues for future work in Appendix \ref{sec:appendix_discussion}.

In closing, the analyses discussed here primarily provide a diagnostic tool to shed light on the intricate interplay between stability and plasticity. They also provide analytical insights for the future development of continual learning algorithms and could potentially also be utilised as a lens to study other desirable properties for continual learners. The journey of continual learning research continues, with our work offering a valuable contribution towards achieving the delicate balance between preserving past knowledge and adapting to new challenges in an ever-evolving world of information.

\section{Reproducibility Statement}

Details about the implementation of the continual learning algorithms are mentioned both in the Sections~\ref{sec:nonlinear} and \ref{sec:linear}, and in the Appendix~\ref{sec:methods}. 

\section{Acknowledgments}
The project was financed by the funds of the research training group “Computational Cognition” (GRK2340) provided by the Deutsche Forschungsgemeinschaft (DFG), Germany and the European Union (ERC, TIME, Project $101039524$). 
Compute resources used for this project are additionally supported by the Deutsche Forschungsgemeinschaft (DFG, German Research Foundation), project number $456666331$.

\bibliography{bibliography}
\bibliographystyle{collas2024_conference}
\appendix

\section{Appendix}

\subsection{Methods}

\label{sec:methods}
\subsubsection{Assessing existing continual learning algorithms with readout-based gradient decomposition}
\label{sec:appendix_nonlinear}

For the experiments on the Cifar110 task, we construct $11$ datasets (one for each task). The first task, on which we perform the bulk of our analyses consists of the full Cifar10 dataset (with usual training and validation splits). For each subsequent task, we sample $10$ unique classes from Cifar100. Experiments are repeated with three different repeats of this procedure, providing some control for the varying difficulty of the different task splits.
Data for all tasks was augmented with random cropping (padding $=4$) and horizontal flipping throughout training. All data was normalized with means ($0.5071$, $0.4865$, $0.4409$) and standard deviations ($0.2673$, $0.2564$, $0.2762$) for the RGB channels.

All networks were trained with Adam ($lr=0.001$, $\beta_1=0.9$, $\beta_2=0.999$) for $60$ epochs per task. 
Following the findings in \cite{li2017learning}, we warm up the new readout at the start of each new task (excluding training on the first task). This has been reported to stabilize representations at the start of training on a new task (where the randomly initialized new readout is not aligned with the features of the remainder of the network, causing large gradients). We freeze the weights of all layers except the new readout for the first $10$ epochs of training.
Additionally, since our analyses investigate activation changes relative to the range of previously learned readouts, we freeze all parameters in old readouts for methods that would otherwise allow changing readout weights for old tasks (this is the case for LwF and data replay).

The network architecture for these experiments is adopted from \citet{zenke2017continual} and has been slightly altered. It consists of two VGG blocks ($32$ channels in the first, $64$ channels in the second block each, kernel size $3$). Each block of two convolutional layers is followed by a max pool layer with kernel size and stride $2$.
The pre-readout dense layer was scaled to have $128$ output units and no dropout was used throughout the network.
All layers in the backbone were initialized with Kaiming-He \cite{he2015delving} initialization as implemented in PyTorch.

After performing initial sweeps for the hyperparameters in the tested algorithms to determine the rough effective ranges, we performed additional sweeps for each algorithm in order to generate the data points in Figure~\ref{fig:nonlinear}. Each data point visualized is the average over three experiments with the same hyperparameter settings, but different seeds (and therefore task splits as described above).

Hyperparameters were swept as follows: 
\begin{itemize}
    \item For EWC, $\lambda$ was varied between $10^{-1}$ - $10^5$.
    \item For SI, $\epsilon$ was fixed to $1$ and $\lambda$ was varied between $10^{-2}$ and $10^5$.
    \item For LwF, we fixed the temperature to $1$ and varied alpha between $0.01$ and $10$.
    \item For data replay, we used replay buffer sizes between $0$ and $60000$ samples, with the default replay buffer style as implemented in Avalanche (as of version $0.3.1$).
    \item For GEM, we varied the memory strength parameter ($\gamma$ in the original publication) between $0$ and $1$ and varied the number of patterns stored per experience to estimate the gradient projection between $0$ and $20000$.
\end{itemize}

\subsubsection{Gradient Decomposition in the linear network}
\label{sec:appendix_linear}

The linear system described in section \ref{sec:linear} is a one-hidden layer network without biases and $11$ units in its hidden layer. 
The network has two separate linear readouts with $5$ units each, to accommodate the split MNIST task. For all experiments, the network was trained for $30$ epochs per task, with plain stochastic gradient descent and a learning rate $5.10^{-4}$ and batch size $16$.

Since the Split MNIST task is very easy, even for a small linear network we increase the difficulty of the dataset slightly by applying a number of transformations to the dataset once at the time of constructing the dataset. 
This increases the effect of catastrophic forgetting while keeping a fixed dataset, allowing for easy experimentation. The transformations were implemented using the torchvision transforms package. Images of digits were augmented with random rotations ($\pm10$ degrees), translations ($\pm10$ percent of image size in both axes), scaled between $90$-$110$\% of the original size and randomly cropped with padding $= 4$. Finally, we applied the 'ColorJitter' transformation with parameters brightness $=0.1,$ contrast $=0.1$, saturation $=0.1$, and hue $=0.1$. Transformations are only applied to training data for both tasks.

For EWC, we approximate the diagonal of the Fisher information matrix for the hidden layer parameters as the square of the gradients for the first task over the whole dataset for task 1.

$$F_w = \frac{\sum_N (\Delta w)^2}{Nb}, $$

for N batches of data (with $b$ samples each). We sweep $1000$ values for the scalar multiplier $\lambda$ governing regularization strength on a log scale between $0$ and $10^5$.

\begin{figure}[t]
\centering
\includegraphics[width=\textwidth]{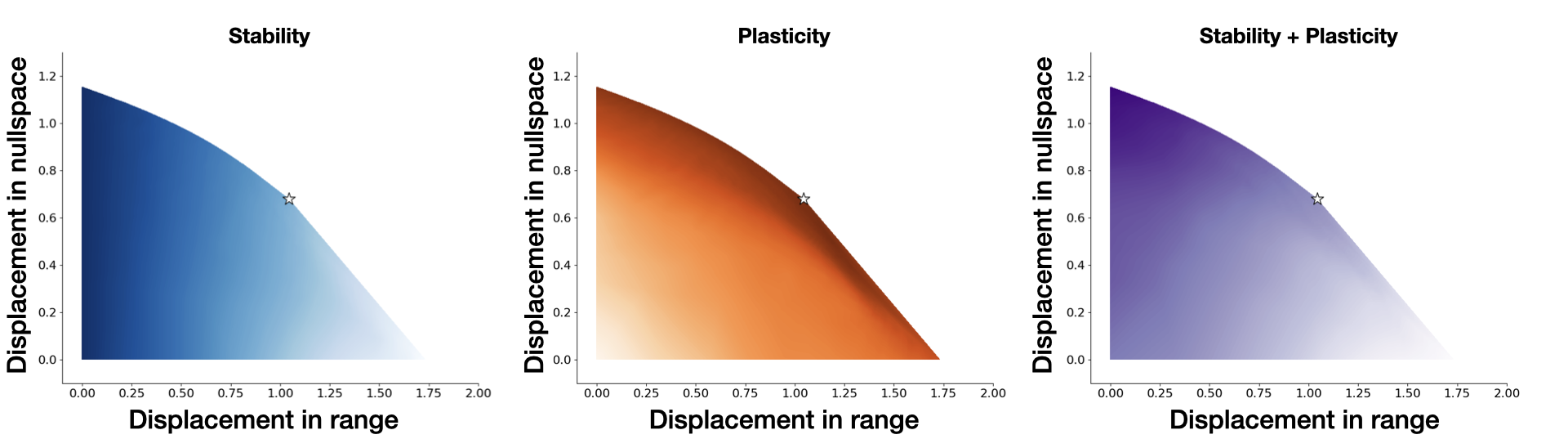}
\caption{\textbf{Movement in range and nullspace for gradient decomposition as discussed in Section \ref{sec:linear}}. The three panels show activation change in range and null space. In each panel the surface is coloured according to a different performance measure analogously to Figure \ref{fig:conceptual}.}
\label{fig:extended_contour}
\end{figure}

To illustrate our gradient decomposition result, we swept the space of possible decompositions in a grid with $33$ linearly spaced values between $0$ and $1$ for $\alpha$ and $\beta$. In Figure~\ref{fig:linear} we visualized the extremes of this search, and the results of the full space are included in Figure~\ref{fig:extended_contour} for completeness.

\subsubsection{Nonlinear approximation of functional range and nullspace}
\label{sec_appendix_nonlinear_analysis}

The deep network described in section \ref{sec:nonlinear} is a VGG-style convolutional network.
As the decomposition of gradient spaces for the weight matrices at each layer is very expensive and scales with the total number of parameters in a layer, all layers are deliberately chosen relatively small. 
Instead, to keep the number of parameters at each layer manageable the model is relatively narrow and deep.
This setup has two additional benefits for our analysis:
First, the presented analysis focuses on the effects of nonlinearities and hierarchy. This relatively deep network is suitable to assess these aspects.
Secondly, in networks with low capacity we expect the stability-plasticity trade-off to be especially pronounced. In cases where the number of dimensions is small, the network needs to be efficient during learning to maintain plasticity. The effects of over constraining the network are expected to be especially visible in the low capacity regime.

The network consists of $5$ VGG blocks consisting of two convolutional layers each. Both convolutional layers use ReLU activation functions and each block ends with a MaxPool operation with 2x2 kernel and stride 2. Convolutional layers within a block share the same number of output channels. The $5$ blocks have $8$, $8$, $16$, $16$, and $32$ channels respectively.
The final hidden layer of the network is a dense layer with $64$ units and ReLU activation function. Each task is initialised with its own readout layer.
All layers in the network including the readout layers are initialised without bias. We find that the network performs well without bias, and omitting it facilitates our analysis.

The network is trained for $30$ epochs each for both tasks. Consistent with the other analyses in this paper the backbone of the model is frozen for the first $5$ epochs of the second task to warm up the new readout.
The network is trained with Adam with default parameters ($lr=0.001, \beta_1=0.9, \beta_2=0.999, \epsilon=1e-8$) and a batchsize of $64$. We use a custom implementation of Adam that allows for projection of the computed updates back into the allowed subspaces. This is necessary because even if the moment estimates are computed on projected gradients, the final update computed based on the gradient of the current batch and the moment estimates can still fall outside of the allowed subspace. We find that the best procedure to ensure the gradients are correctly projected is to first project the true gradients computed for a batch, then use the projected gradients to compute the weight update with Adam, and finally re-project the weight update using the same projection matrices. We re-initialise Adam for the second task to ensure the final moment estimates from the first task do not affect learning for the second task.

Contrary to the subspaces estimated based on the readout weight matrix the subspace estimates based on old task gradients used here are not constant. The reason for this is as follows: The subspaces are linear estimates based on the gradients for the old task computed at a specific location in weight space. Even if all weight updates are orthogonal to the space spanned by these gradients the non-convex nature of the loss landscape can lead to a decay in accuracy of the estimated subspaces. Therefore, in the diagnostic setting, it is beneficial to frequently re-estimate the subspaces.
During learning for the second task, we re-estimate the subspaces every 200 gradient descent steps. Every time the update is computed, we compute gradients for 20 epochs of data from the training set of the old task. This ensures that the resulting matrix has more gradient samples than there are parameters in the largest layer of the network. For each matrix of gradient estimates we perform SVD and keep the singular vectors whose singular values together explain 99.9\% variance as the orthonormal basis of the functional range at this layer.

We construct the Split CIFAR-10 dataset from the CIFAR-10 dataset available in torchvision, with the usual training and validation splits.
For each of the three seeds we test for each parameter configuration the dataset is split into two tasks with 5 randomly selected classes each (sampled without replacement). No transformations are applied to the dataset.

As an additional point of comparison we train the same network for varying regularisation strength using EWC. EWC regularisation strengths are swept between $0.001$ - $10000000$ (regularisation strength increases one order of magnitude between each run, 10 configurations total each repeated for $3$ seeds).

\begin{figure}[t]
    \centering
    \includegraphics[width=\textwidth]{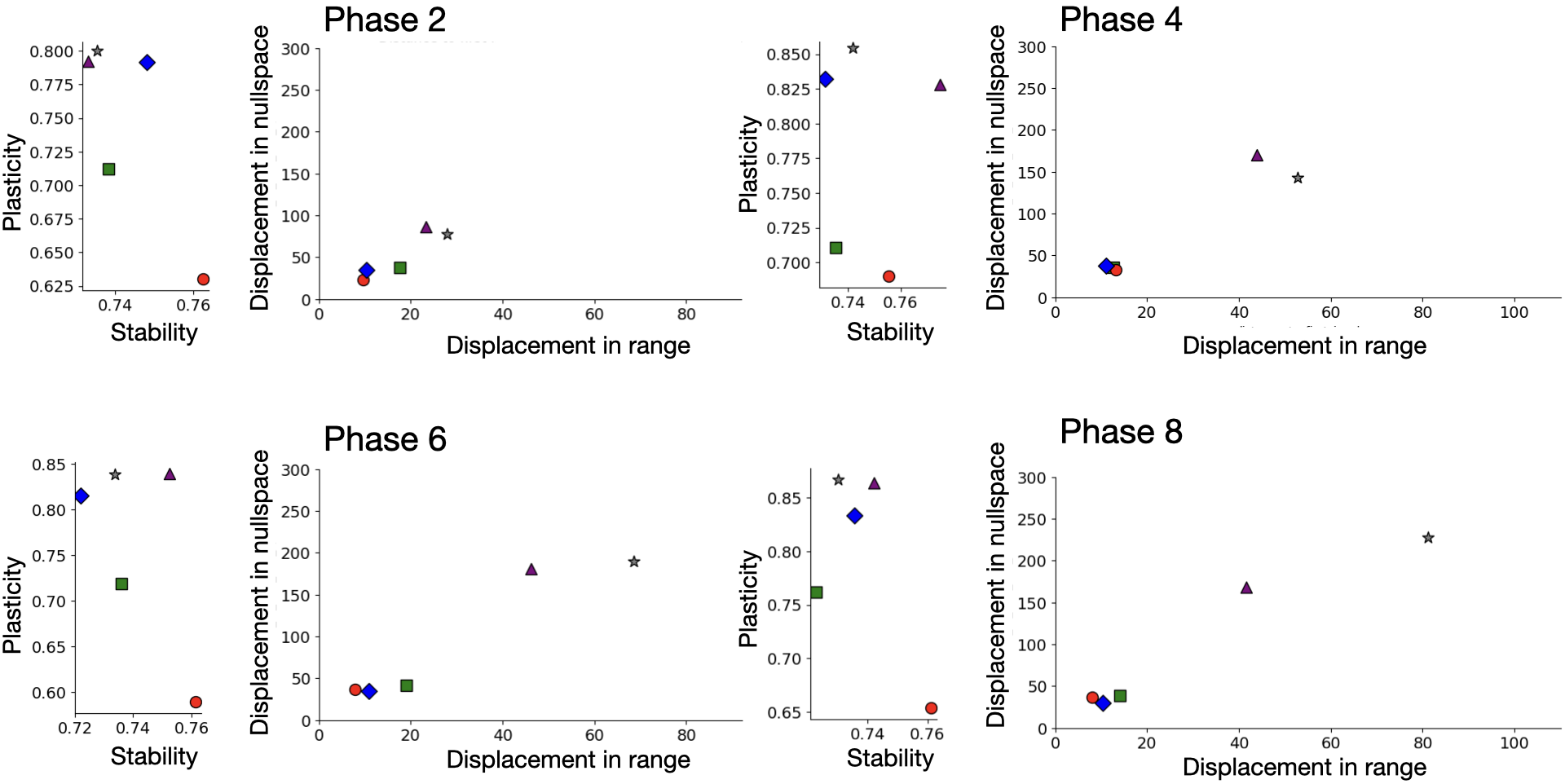}
    \caption{\textbf{Behavior and analysis of the continual learning algorithms from Figure \ref{fig:comparison} across the training phases.} For each phase of training, we select one regularization strength setting for each algorithm such that the stability is very high and relatively similar (here $\sim$0.735). We monitor the plasticity, and the displacements in range and null space of the readout for task 1, across training phases, to assess how quickly the stability-plasticity trade-offs of the algorithms diverge. We observe that the stability-plasticity trade-offs observed in Figure \ref{fig:comparison} already manifest in Phase 2 i.e. after training on two additional tasks.}
    \label{fig:comparison_over_phases}
\end{figure}

\subsection{Assessment of continual learning on a larger network and dataset}
\label{sec:appendix_tinyimg}

\begin{figure}[]
\centering
\includegraphics[width=\textwidth]{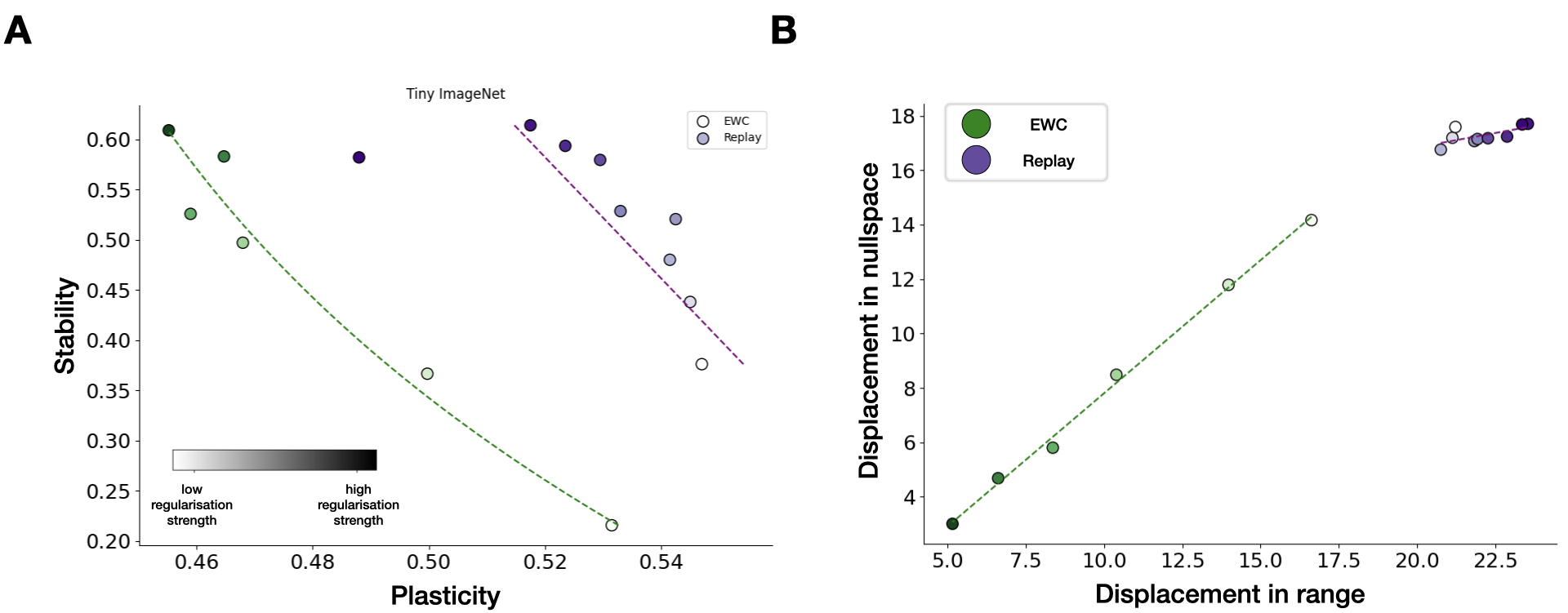}
\caption{\textbf{Stability-Plasticity tradeoff as observed after training on 5 splits of the TinyImagenet dataset.} \textbf{(A)} Displacement of representations of the first task after training on all 5 tasks, decomposed into nullspace and range movement using our readout decomposition method. \textbf{(B)} Analogous to the results observed in Cifar110, we see that replay can achieve high stability while sacrificing less plasticity compared to the regularisation-based method EWC. As observed before, higher plasticity observed in replay correlates with more movement of previous task representations. EWC strongly restricts the movement of previously learned representations and cannot maintain high plasticity as regularisation strength increases.}
\label{fig:tinyimg}
\end{figure}

To assess whether our findings on the Cifar110 scale to larger tasks and stimuli, we repeat our analysis of EWC and data replay (previous readouts frozen) on the TinyImagenet dataset \cite{le2015tiny}, adapting the slimmed ResNet18 as reported in \cite{lopez2017gradient} and implemented in Avalanche \cite{carta2023avalanche}. TinyImagenet consists of 64x64x3 images belonging to 200 classes, which we split into 5 unique subsets of 40 classes per task.

As before, we sweep over regularisation strengths for both EWC and Replay. For EWC, we sweep lambda in [1 .. 100000], for Replay we vary replay buffer sizes in [1000 .. 50000]. To allow for estimation of movement in the nullspace and range of previous readouts, we freeze all parameters in the readout layers for previously trained tasks.

All networks were trained for 60 epochs per task, using the Adam optimizer with the same settings as used for our earlier experiment. For tasks 2 to 5, we only train the new readout for the first 10 epochs, to align the new readout with the rest of the network before propagating gradients.

\subsection{Continual learning in a three-layer linear neural network}
\label{sec:appendix_3layer}
In order to demonstrate the complexity of deriving an efficient gradient decomposition algorithm (cf. Section~\ref{sec:linear}) for multi-layer linear networks, we consider the case of a three-hidden layer linear network: $\mathbf{o = xW_{H_1}^{\top}W_{H_2}^{\top}W_{H_3}^{\top}W_R^{\top}}$.  

After training on task $1$: $\{\mathbf{x^1\rightarrow o^1}\}$, we get the trained readout $\mathbf{W_{R^1}}$. While training on task $2$: $\{\mathbf{x^2\rightarrow o^2}\}$, we get the gradient $\mathbf{\Delta\!W_H}$. In order to maintain stability, we want the learned task $1$: $\{\mathbf{x^1\rightarrow o^1}\}$ mapping to stay preserved. 

We want: $\mathbf{o^1 = x^1(W_{H_1}+\Delta\!W_{H_1})^{\top}(W_{H_2}+\Delta\!W_{H_2})^{\top}(W_{H_3}+\Delta\!W_{H_3})^{\top}W_{R^1}^{\top}}$, implying:
\begin{align}
\notag&\mathbf{x^1(\Delta\!W_{H_1}^{\top}W_{H_2}^{\top}W_{H_3}^{\top} + W_{H_1}^{\top}\Delta\!W_{H_2}^{\top}W_{H_3}^{\top} + W_{H_1}^{\top}W_{H_2}^{\top}\Delta\!W_{H_3}^{\top} + W_{H_1}^{\top}\Delta\!W_{H_2}^{\top}\Delta\!W_{H_3}^{\top}} + \\
&\mathbf{\Delta\!W_{H_1}^{\top}W_{H_2}^{\top}\Delta\!W_{H_3}^{\top} + \Delta\!W_{H_1}^{\top}\Delta\!W_{H_2}^{\top}W_{H_3}^{\top} + \Delta\!W_{H_1}^{\top}\Delta\!W_{H_2}^{\top}\Delta\!W_{H_3}^{\top})W_{R^1}^{\top} = 0}\label{eq:3layer}
\end{align}

There could be multiple ways of constraining the gradients to satisfy Eq.~\ref{eq:3layer}. However, we explored if, parsimoniously, a solution could only constrain the pre-readout gradient $\Delta\!W_{H_3}$ and let backpropagation take care of the rest.

Using backpropagation we can write out the gradients in terms of the derivative of the loss function, $\mathbf{e_o} = \frac{\partial \mathcal{L}(\mathbf{o,o^2})}{\partial \mathbf{o}}$ (assuming batch size $1$ here, mapping $\mathbf{x^2\rightarrow o^2}$). $\mathbf{e_o}$, and not $\Delta\!W_{H_3}$, is propagated back for upstream gradient computations as it is independent of the network activations. The gradients are computed as follows:
\begin{align*}
    \mathbf{\Delta\!W_{H_1}^{\top}} &\mathbf{= (x^2)^{\top}e_o W_{R^2}W_{H_3}W_{H_2} }\\
    \mathbf{\Delta\!W_{H_2}^{\top}} &\mathbf{= W_{H_1}(x^2)^{\top}e_o W_{R^2}W_{H_3}} \\
    \mathbf{\Delta\!W_{H_3}^{\top}} &\mathbf{= W_{H_2}W_{H_1}(x^2)^{\top}e_o W_{R^2}}
\end{align*}

We would like to know the transformation $\mathbf{e_o\rightarrow Ae_o}$ which satisfies the constraint in Eq.~\ref{eq:3layer}, when this transformed $\mathbf{e_o}$ is backpropagated. As a first step, we can ignore the interactions between the terms of Eq.~\ref{eq:3layer} by asking them to be independently $0$. The resulting transformation is jointly subject to $3$ constraints:
\begin{enumerate}
    \item In order to maintain a non-zero gradient $\mathbf{\Delta\!W_{H_3}^{\top}}$, while zeroing the terms associated with it in Eq.~\ref{eq:3layer}, $\mathbf{A}$ should project gradients into the nullspace of $\mathbf{W_{R^1}W_{R^2}^{\top}}$
    \item In order to maintain a non-zero gradient $\mathbf{\Delta\!W_{H_2}^{\top}}$, while zeroing the remaining terms associated with it in Eq.~\ref{eq:3layer}, $\mathbf{A}$ should also project gradients into the nullspace of $\mathbf{W_{R^1}W_{H_3}W_{H_3}^{\top}W_{R^2}^{\top}}$
    \item In order to maintain a non-zero gradient $\mathbf{\Delta\!W_{H_1}^{\top}}$, while zeroing the remaining term associated with it in Eq.~\ref{eq:3layer}, $\mathbf{A}$ should also project gradients into the nullspace of $\mathbf{W_{R^1}W_{H_3}W_{H_2}W_{H_2}^{\top}W_{H_3}^{\top}W_{R^2}^{\top}}$
\end{enumerate}

Intuitively, this is similar to canceling the propagation of $\mathbf{e_o}$ into readout $1$ (through all of the $3$ paths listed above) i.e. any errors that would be induced in readout $1$ by changing any of the weights would become zero due to such a projection, $\mathbf{Ae_o}$. This algorithm would ensure stability, however, it is unclear how much plasticity can be leveraged after the said projection - if the intersection of the nullspaces spans a very low-dimensional space, plasticity will be hampered. This needs to be studied empirically for a variety of datasets.

This parsimonious algorithm is computationally expensive as compared to the gradient decomposition algorithm for the one-hidden layer network discussed in Section~\ref{sec:linear}. In the current algorithm, during every weight update, the $3$ nullspaces need to be computed, as the weights keep changing. Additionally, the number of nullspaces to be computed scales with the number of hidden layers $n$ and with the number of tasks $k$, as $n(k-1)$. With increasing $n$ and $k$, the intersections of the nullspaces would get smaller and plasticity would be hampered. How much of this is a problem for existing continual learning datasets needs to be tested empirically.

\subsection{Discussion on the generality of our findings and future steps}
\label{sec:appendix_discussion}

The link between the stability-plasticity trade-off and activation changes in the range and null space of readouts can be extended beyond task-incremental learning to other continual learning settings - class-incremental and domain-incremental learning~\citep{van2022three}. The core intuition is that the subspaces relevant for the projection of the pre-readout activations, corresponding to prior classes or domains, to the output, must be preserved to maintain stability. In learning newer classes or domains, learning should not be restricted in the remainder of the subspaces available in the projection of the pre-readout activations to the output, to maintain plasticity.

Additionally, the framework we discuss here can potentially be useful to study other properties of continual learners as well: One important aspect by which continual learners can be evaluated is how quickly they can learn to perform additional tasks. We hypothesise that speed of learning may be correlated with the size of the subspaces in a network in which learning is not restricted in service of stability. In line with our conclusions on achieving optimal plasticity, we expect that limiting learning only in the functional ranges of previous tasks should lead to fastest learning in future tasks. A second interesting question for future research is whether preserving only features that lie in the functional ranges of previous tasks will affect the likelihood of forward transfer.  We hypothesise that those features that were functionally relevant for previous tasks are most likely to be relevant for future tasks, especially in scenarios where all tasks share relatively similar dataset statistics (such as naturalistic stimuli). Preserving these features while allowing as much space as possible for new features to be added in new tasks could likely facilitate forward transfer. Experimental treatment of these hypotheses is a promising avenue for future research.
\end{document}

%% file: math_commands.tex
\usepackage{amsmath,amsfonts,bm}

\def\eqref#1{equation~\ref{#1}}

\def\1{\bm{1}}

\DeclareMathAlphabet{\mathsfit}{\encodingdefault}{\sfdefault}{m}{sl}
\SetMathAlphabet{\mathsfit}{bold}{\encodingdefault}{\sfdefault}{bx}{n}













%% file: main.bbl
\begin{thebibliography}{35}
\providecommand{\natexlab}[1]{#1}
\providecommand{\url}[1]{\texttt{#1}}
\expandafter\ifx\csname urlstyle\endcsname\relax
  \providecommand{\doi}[1]{doi: #1}\else
  \providecommand{\doi}{doi: \begingroup \urlstyle{rm}\Url}\fi

\bibitem[Anthes et~al.(2023)Anthes, Thorat, König, and
  Kietzmann]{anthes2023diagnosing}
Daniel Anthes, Sushrut Thorat, Peter König, and Tim~C Kietzmann.
\newblock Diagnosing catastrophe: Large parts of accuracy loss in continual
  learning can be accounted for by readout misalignment.
\newblock In \emph{Conference on Cognitive Computational Neuroscience}, pp.\
  748--751, 2023.

\bibitem[Bagus \& Gepperth(2021)Bagus and Gepperth]{bagus2021investigation}
Benedikt Bagus and Alexander Gepperth.
\newblock An investigation of replay-based approaches for continual learning.
\newblock In \emph{2021 International Joint Conference on Neural Networks
  (IJCNN)}, pp.\  1--9. IEEE, 2021.

\bibitem[Carpenter \& Grossberg(1987)Carpenter and
  Grossberg]{carpenter1987massively}
Gail~A Carpenter and Stephen Grossberg.
\newblock A massively parallel architecture for a self-organizing neural
  pattern recognition machine.
\newblock \emph{Computer vision, graphics, and image processing}, 37\penalty0
  (1):\penalty0 54--115, 1987.

\bibitem[Carta et~al.(2023)Carta, Pellegrini, Cossu, Hemati, and
  Lomonaco]{carta2023avalanche}
Antonio Carta, Lorenzo Pellegrini, Andrea Cossu, Hamed Hemati, and Vincenzo
  Lomonaco.
\newblock Avalanche: A pytorch library for deep continual learning.
\newblock \emph{Journal of Machine Learning Research}, 24\penalty0
  (363):\penalty0 1--6, 2023.

\bibitem[Davari \& Belilovsky(2021)Davari and Belilovsky]{davari2021probing}
Mohammad~Reza Davari and Eugene Belilovsky.
\newblock Probing representation forgetting in continual learning.
\newblock In \emph{NeurIPS 2021 Workshop on Distribution Shifts: Connecting
  Methods and Applications}, 2021.

\bibitem[De~Lange et~al.(2021)De~Lange, Aljundi, Masana, Parisot, Jia,
  Leonardis, Slabaugh, and Tuytelaars]{de2021continual}
Matthias De~Lange, Rahaf Aljundi, Marc Masana, Sarah Parisot, Xu~Jia,
  Ale{\v{s}} Leonardis, Gregory Slabaugh, and Tinne Tuytelaars.
\newblock A continual learning survey: Defying forgetting in classification
  tasks.
\newblock \emph{IEEE transactions on pattern analysis and machine
  intelligence}, 44\penalty0 (7):\penalty0 3366--3385, 2021.

\bibitem[Deng et~al.(2021)Deng, Chen, Hao, Wang, and Heng]{deng2021flattening}
Danruo Deng, Guangyong Chen, Jianye Hao, Qiong Wang, and Pheng-Ann Heng.
\newblock Flattening sharpness for dynamic gradient projection memory benefits
  continual learning.
\newblock \emph{Advances in Neural Information Processing Systems},
  34:\penalty0 18710--18721, 2021.

\bibitem[Dohare et~al.(2023)Dohare, Hernandez-Garcia, Rahman, Sutton, and
  Mahmood]{dohare2023loss}
Shibhansh Dohare, Juan~Fernando Hernandez-Garcia, Parash Rahman, Richard~S.
  Sutton, and A.~Rupam Mahmood.
\newblock Loss of plasticity in deep continual learning.
\newblock \emph{Research Square preprint}, 2023.

\bibitem[Farajtabar et~al.(2020)Farajtabar, Azizan, Mott, and
  Li]{farajtabar2020orthogonal}
Mehrdad Farajtabar, Navid Azizan, Alex Mott, and Ang Li.
\newblock Orthogonal gradient descent for continual learning.
\newblock In \emph{International Conference on Artificial Intelligence and
  Statistics}, pp.\  3762--3773. PMLR, 2020.

\bibitem[French(1999)]{french1999catastrophic}
Robert~M French.
\newblock Catastrophic forgetting in connectionist networks.
\newblock \emph{Trends in cognitive sciences}, 3\penalty0 (4):\penalty0
  128--135, 1999.

\bibitem[Hadsell et~al.(2020)Hadsell, Rao, Rusu, and
  Pascanu]{hadsell2020embracing}
Raia Hadsell, Dushyant Rao, Andrei~A Rusu, and Razvan Pascanu.
\newblock Embracing change: Continual learning in deep neural networks.
\newblock \emph{Trends in cognitive sciences}, 24\penalty0 (12):\penalty0
  1028--1040, 2020.

\bibitem[He et~al.(2015)He, Zhang, Ren, and Sun]{he2015delving}
Kaiming He, Xiangyu Zhang, Shaoqing Ren, and Jian Sun.
\newblock Delving deep into rectifiers: Surpassing human-level performance on
  imagenet classification.
\newblock In \emph{Proceedings of the IEEE international conference on computer
  vision}, pp.\  1026--1034, 2015.

\bibitem[He(2018)]{he2018continual}
Xu~He.
\newblock Continual learning by conceptor regularization.
\newblock In \emph{Continual learning workshop (NeurIPS 2018). url:
  https://sites. google. com/view/continual2018/submissions}, 2018.

\bibitem[He \& Jaeger(2017)He and Jaeger]{he2017overcoming}
Xu~He and Herbert Jaeger.
\newblock Overcoming catastrophic interference by conceptors.
\newblock \emph{arXiv preprint arXiv:1707.04853}, 2017.

\bibitem[Hong et~al.(2016)Hong, Yamins, Majaj, and DiCarlo]{hong2016explicit}
Ha~Hong, Daniel~LK Yamins, Najib~J Majaj, and James~J DiCarlo.
\newblock Explicit information for category-orthogonal object properties
  increases along the ventral stream.
\newblock \emph{Nature neuroscience}, 19\penalty0 (4):\penalty0 613--622, 2016.

\bibitem[Kalb \& Beyerer(2022)Kalb and Beyerer]{kalb2022causes}
Tobias Kalb and J{\"u}rgen Beyerer.
\newblock Causes of catastrophic forgetting in class-incremental semantic
  segmentation.
\newblock In \emph{Proceedings of the Asian Conference on Computer Vision},
  pp.\  56--73, 2022.

\bibitem[Kirkpatrick et~al.(2017)Kirkpatrick, Pascanu, Rabinowitz, Veness,
  Desjardins, Rusu, Milan, Quan, Ramalho, Grabska-Barwinska,
  et~al.]{kirkpatrick2017overcoming}
James Kirkpatrick, Razvan Pascanu, Neil Rabinowitz, Joel Veness, Guillaume
  Desjardins, Andrei~A Rusu, Kieran Milan, John Quan, Tiago Ramalho, Agnieszka
  Grabska-Barwinska, et~al.
\newblock Overcoming catastrophic forgetting in neural networks.
\newblock \emph{Proceedings of the national academy of sciences}, 114\penalty0
  (13):\penalty0 3521--3526, 2017.

\bibitem[Kong et~al.(2022)Kong, Liu, Wang, and Tao]{kong2022balancing}
Yajing Kong, Liu Liu, Zhen Wang, and Dacheng Tao.
\newblock Balancing stability and plasticity through advanced null space in
  continual learning.
\newblock In \emph{European Conference on Computer Vision}, pp.\  219--236.
  Springer, 2022.

\bibitem[Krizhevsky et~al.(2009)Krizhevsky, Hinton,
  et~al.]{krizhevsky2009learning}
Alex Krizhevsky, Geoffrey Hinton, et~al.
\newblock Learning multiple layers of features from tiny images.
\newblock 2009.

\bibitem[Le \& Yang(2015)Le and Yang]{le2015tiny}
Ya~Le and Xuan Yang.
\newblock Tiny imagenet visual recognition challenge.
\newblock \emph{CS 231N}, 7\penalty0 (7):\penalty0 3, 2015.

\bibitem[Li \& Hoiem(2017)Li and Hoiem]{li2017learning}
Zhizhong Li and Derek Hoiem.
\newblock Learning without forgetting.
\newblock \emph{IEEE transactions on pattern analysis and machine
  intelligence}, 40\penalty0 (12):\penalty0 2935--2947, 2017.

\bibitem[Lopez-Paz \& Ranzato(2017)Lopez-Paz and Ranzato]{lopez2017gradient}
David Lopez-Paz and Marc'Aurelio Ranzato.
\newblock Gradient episodic memory for continual learning.
\newblock \emph{Advances in neural information processing systems}, 30, 2017.

\bibitem[McCloskey \& Cohen(1989)McCloskey and
  Cohen]{mccloskey1989catastrophic}
Michael McCloskey and Neal~J Cohen.
\newblock Catastrophic interference in connectionist networks: The sequential
  learning problem.
\newblock In \emph{Psychology of learning and motivation}, volume~24, pp.\
  109--165. Elsevier, 1989.

\bibitem[Mermillod et~al.(2013)Mermillod, Bugaiska, and
  Bonin]{mermillod2013stability}
Martial Mermillod, Aur{\'e}lia Bugaiska, and Patrick Bonin.
\newblock The stability-plasticity dilemma: Investigating the continuum from
  catastrophic forgetting to age-limited learning effects.
\newblock \emph{Frontiers in psychology}, 4:\penalty0 504, 2013.

\bibitem[Parisi et~al.(2019)Parisi, Kemker, Part, Kanan, and
  Wermter]{parisi2019continual}
German~I Parisi, Ronald Kemker, Jose~L Part, Christopher Kanan, and Stefan
  Wermter.
\newblock Continual lifelong learning with neural networks: A review.
\newblock \emph{Neural networks}, 113:\penalty0 54--71, 2019.

\bibitem[Ramasesh et~al.(2020)Ramasesh, Dyer, and Raghu]{ramasesh2020anatomy}
Vinay~V Ramasesh, Ethan Dyer, and Maithra Raghu.
\newblock Anatomy of catastrophic forgetting: Hidden representations and task
  semantics.
\newblock \emph{arXiv preprint arXiv:2007.07400}, 2020.

\bibitem[Ramasesh et~al.(2021)Ramasesh, Lewkowycz, and
  Dyer]{ramasesh2021effect}
Vinay~Venkatesh Ramasesh, Aitor Lewkowycz, and Ethan Dyer.
\newblock Effect of scale on catastrophic forgetting in neural networks.
\newblock In \emph{International Conference on Learning Representations}, 2021.

\bibitem[Rebuffi et~al.(2017)Rebuffi, Kolesnikov, Sperl, and
  Lampert]{rebuffi2017icarl}
Sylvestre-Alvise Rebuffi, Alexander Kolesnikov, Georg Sperl, and Christoph~H
  Lampert.
\newblock icarl: Incremental classifier and representation learning.
\newblock In \emph{Proceedings of the IEEE conference on Computer Vision and
  Pattern Recognition}, pp.\  2001--2010, 2017.

\bibitem[Saha et~al.(2021)Saha, Garg, and Roy]{saha2021gradient}
Gobinda Saha, Isha Garg, and Kaushik Roy.
\newblock Gradient projection memory for continual learning.
\newblock \emph{arXiv preprint arXiv:2103.09762}, 2021.

\bibitem[Thorat et~al.(2021)Thorat, Aldegheri, and
  Kietzmann]{thorat2021categoryorthogonal}
Sushrut Thorat, Giacomo Aldegheri, and Tim~C Kietzmann.
\newblock Category-orthogonal object features guide information processing in
  recurrent neural networks trained for object categorization.
\newblock In \emph{SVRHM 2021 Workshop @ NeurIPS}, 2021.

\bibitem[van~de Ven et~al.(2022)van~de Ven, Tuytelaars, and
  Tolias]{van2022three}
Gido~M van~de Ven, Tinne Tuytelaars, and Andreas~S Tolias.
\newblock Three types of incremental learning.
\newblock \emph{Nature Machine Intelligence}, 4\penalty0 (12):\penalty0
  1185--1197, 2022.

\bibitem[Wang et~al.(2021)Wang, Li, Sun, and Xu]{wang2021training}
Shipeng Wang, Xiaorong Li, Jian Sun, and Zongben Xu.
\newblock Training networks in null space of feature covariance for continual
  learning.
\newblock In \emph{Proceedings of the IEEE/CVF conference on Computer Vision
  and Pattern Recognition}, pp.\  184--193, 2021.

\bibitem[Zenke et~al.(2017)Zenke, Poole, and Ganguli]{zenke2017continual}
Friedemann Zenke, Ben Poole, and Surya Ganguli.
\newblock Continual learning through synaptic intelligence.
\newblock In \emph{International conference on machine learning}, pp.\
  3987--3995. PMLR, 2017.

\bibitem[Zhang et~al.(2021)Zhang, Bengio, Hardt, Recht, and
  Vinyals]{zhang2021understanding}
Chiyuan Zhang, Samy Bengio, Moritz Hardt, Benjamin Recht, and Oriol Vinyals.
\newblock Understanding deep learning (still) requires rethinking
  generalization.
\newblock \emph{Communications of the ACM}, 64\penalty0 (3):\penalty0 107--115,
  2021.

\bibitem[Zhao et~al.(2023)Zhao, Zhang, Tan, Liu, Qu, Xie, and
  Ma]{zhao2023rethinking}
Zhen Zhao, Zhizhong Zhang, Xin Tan, Jun Liu, Yanyun Qu, Yuan Xie, and Lizhuang
  Ma.
\newblock Rethinking gradient projection continual learning:
  Stability/plasticity feature space decoupling.
\newblock In \emph{Proceedings of the IEEE/CVF Conference on Computer Vision
  and Pattern Recognition}, pp.\  3718--3727, 2023.

\end{thebibliography}
